\documentclass[lettersize,journal]{IEEEtran}
\usepackage{amsmath,amsfonts}

\usepackage{array}
\usepackage[caption=false,font=normalsize,labelfont=sf,textfont=sf]{subfig}
\usepackage{textcomp}
\usepackage{stfloats}
\usepackage{url}
\usepackage{verbatim}
\usepackage{graphicx}
\usepackage{cite}

\usepackage[utf8]{inputenc}
\usepackage{amsmath}
\usepackage{algorithm}
\usepackage{algpseudocode}
\usepackage{hyperref}

\usepackage{multirow}
\usepackage{listings}
\usepackage{color}
\usepackage{xcolor}
\usepackage{colortbl}
\definecolor{dkgreen}{rgb}{0,0.6,0}
\definecolor{gray}{rgb}{0.5,0.5,0.5}
\definecolor{mauve}{rgb}{0.58,0,0.82}
\begin{document}
	
\title{Uncertainty Guided Refinement for Fine-Grained Salient Object Detection}

\author{Yao Yuan,
        Pan Gao,
        Qun Dai,
        Jie Qin,
        and Wei Xiang
    \thanks{This work was supported in part by  the Natural Science Foundation of China under Grant 62272227 and Grant 62276129, and in part by High  Performance  Computing  Platform  of Nanjing University of Aeronautics and Astronautics. \emph{(Corresponding author: Pan Gao; Jie Qin.)}}
	\thanks{Y. Yuan, P. Gao, Q. Dai, and J. Qin are with Collge of Computer Science and Technology, Nanjing University of Aeronautics and Astronautics, and also with the Key Laboratory of Brain-Machine Intelligence
    Technology, Ministry of Education, Nanjing 211106, China (e-mail: \{ayews233, pan.gao, daiqun, jie.qin\}@nuaa.edu.cn).}
    \thanks{W. Xiang is with the School of Computing, Engineering and Mathematical Sciences, La Trobe University, Melbourne VIC 3086, Australia  (e-mail: w.xiang@latrobe.edu.au).}
	}

\markboth{Journal of \LaTeX\ Class Files,~Vol.~14, No.~8, August~2021}%
{Shell \MakeLowercase{\textit{et al.}}: A Sample Article Using IEEEtran.cls for IEEE Journals}


\maketitle
	
\begin{abstract}
Recently, salient object detection (SOD) methods have achieved impressive performance. 
However, salient regions predicted by existing methods usually contain unsaturated regions and shadows, 
which limits the model for reliable fine-grained predictions. 
To address this, we introduce the uncertainty guidance learning approach to SOD, 
intended to enhance the model's perception of uncertain regions. 
Specifically, we design a novel Uncertainty Guided Refinement Attention Network (UGRAN), 
which incorporates three important components, 
i.e., the Multilevel Interaction Attention (MIA) module, 
the Scale Spatial-Consistent Attention (SSCA) module, and the Uncertainty Refinement Attention (URA) module. 
Unlike conventional methods dedicated to enhancing features, 
the proposed MIA facilitates the interaction and perception of multilevel features, 
leveraging the complementary characteristics among multilevel features. 
Then, through the proposed SSCA, the salient information across diverse scales 
within the aggregated features can be integrated more comprehensively and integrally. 
In the subsequent steps, we utilize the uncertainty map generated from the saliency prediction map 
to enhance the model's perception capability of uncertain regions, 
generating a highly-saturated fine-grained saliency prediction map. 
Additionally, we devise an adaptive dynamic partition (ADP) mechanism 
to minimize the computational overhead of the URA module 
and improve the utilization of uncertainty guidance. 
Experiments on seven benchmark datasets demonstrate the superiority of 
the proposed UGRAN over the state-of-the-art methodologies. 
Codes will be released at \href{https://github.com/I2-Multimedia-Lab/UGRAN} {https://github.com/I2-Multimedia-Lab/UGRAN}.
\end{abstract}


\section{Introduction}
\IEEEPARstart{S}{alient} Object Detection\cite{SalObjBenchmark} (SOD) aims to identify the most visually attractive objects or regions in an image. 
As SOD is widely applied in the domain of computer vision, it plays a crucial role in many downstream tasks, 
including object detection\cite{article}, semantic segmentation \cite{sun2020mining}, 
image retargeting \cite{8002595, avidan2007seam}, video summarization \cite{4587842}, etc. 

Convolutional Neural Network (CNN) based methods commonly employ encoder-decoder architectures, such as the U-shape~\cite{Unet} based structures, to reconstruct high-quality feature maps by using their multilevel features. This reconstruction can be done either at the encoder side~\cite{8049485,wu2022salient,Amulet,8578428,8578285,Wang_2019_CVPR,BASNet,CPD} or the decoder side~\cite{9889093,9224166,CPD,PAGRN,DSS,DHSNet,wang2023pixels}. Recently, inspired by the success of the Visual Transformer (ViT)~\cite{Vit} in image classification, there have been studies that apply Transformer architectures to dense prediction tasks such as semantic segmentation~\cite{SETR,xie2021segformer,Zhang_2022_CVPR, cheng2023hybrid, zhang2023weakly} or SOD~\cite{VST,SelfReformer,BBRF}. These Transformer-based SOD methods~\cite{VST,SelfReformer,BBRF}, which employ a hierarchical architecture similar to those CNN models, also directly aggregate multilevel features to achieve high-quality saliency predictions. 

\begin{figure}
	\centering
	\includegraphics[scale=0.88]{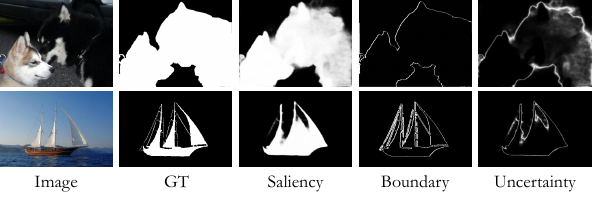}
	\vspace{-2em}\caption{The uncertainty map, compared to the boundary map, 
		more accurately reflects the areas with artifacts and low saturation in the current 
		saliency prediction, thus providing more targeted guidance for the model. }
	\label{fig:motivation}
\end{figure}

Existing methods have achieved promising results in accurately locating salient objects. 
However, fine-grained prediction near object boundaries still poses challenges. 
As shown in Fig. \ref{fig:motivation}, current methods generate 
prediction maps that frequently contain artifacts or unsaturated regions, 
significantly undermining the reliability of the prediction maps. 
To meet this challenge, most existing methods leverage the aggregation of 
low-level features to achieve fine-grained localization of local details, 
since low-level features contain rich local semantics. 
For example, DSS~\cite{DSS} introduced short connections between shallower and deeper layer outputs 
to highlight the entire salient object and accurately locate its boundary. 
NLDF~\cite{luo2017non} extracted the local and global information and combined them via a multi-resolution grid structure. 

Another key mechanism for fine-grained prediction is boundary guidance. 
Due to the presence of blurriness and unsaturated areas in saliency prediction maps, 
typically near object boundaries, 
boundary information is considered competent to provide effective guidance. 
Specifically, EGNet \cite{EGNet} extracts salient object features and salient boundary features separately, 
utilizing their complementarity to alleviate blurred prediction. 
PoolNet~\cite{PoolNet} utilize joint training with boundary detection to
improve the model's ability to perceive boundary regions. 
Besides, F$^3$Net\cite{F3Net} assigned larger weights to boundary pixels in the loss functions, 
enabling the model to focus more on boundary regions. 

Despite the impressive performance, these methods may suffer from the following weaknesses: 
(1) Consider the presence of a substantial amount of noise and interference in low-level features, 
as well as the feature information possibly from non-salient objects, 
the benefit brought by the aggregation of low-level features is limited. 
Only a small fraction of meaningful information is available for fine-grained prediction near object boundaries. 
(2) As shown in Fig. \ref{fig:motivation}, 
boundary guidance based on prior knowledge provided during all stages of training and inference remains fixed, 
which cannot be adapted to the model prediction of where the actual unsaturated region is located. 

To overcome these, this paper proposes a framework dubbed Uncertainty Guided Refinement Attention Network (UGRAN), 
that incorporates the Multilevel Interaction Attention (MIA) module, the Scale Spatial-Consistent Attention (SSCA) module, 
and the Uncertainty Refinement Attention (URA) module with an Adaptive Dynamic Partition (ADP) mechanism. 

First, we rethink how to make full use of multilevel features extracted by the backbone. 
After an in-depth study of the decoder network structures in existing methods, we have drawn two conclusions: 
(1) The process of enhancing multilevel features needs to be redesigned by incorporating the characteristics of features at different scales. 
(2) When integrating the salient information within the aggregated features, the effect of spatial inconsistency of salient information needs to be considered. 
Based on these findings, we propose the MIA module, which can facilitate interaction and perception 
among features at different levels and {reduce non-salient noise within low-level features} via high-to-low interaction attention. 
Then, we introduce the SSCA module to integrate salient information at different scales within the aggregated features. 
{By downsampling the aggregated features to a low resolution, 
the proposed SSCA can enable more effective capture of salient representations at a global level, thereby enhancing the precision of salient object localization. }
These two modules collaborate and complement each other. 

Second, we introduce an uncertainty guidance learning approach. 
As shown in Fig. \ref{fig:motivation}, the uncertainty map is generated based on the current prediction map, 
and can more effectively identify and highlight regions within the saliency map that 
exhibit undersaturation or contain artifacts. Thus, we propose an Uncertainty Refinement Attention (URA) module, 
which utilizes the uncertainty map to guide the refinement of the saliency map. 
Specifically, we employ the uncertainty map as a mask to compute masked attention between the refined feature and the low-level feature, 
enabling the model to prioritize perceiving regions of uncertainty. 
Compared to existing methods, we offer a more explicit approach of
guided learning, which effectively enhances the quality of saliency prediction. 

\begin{figure}
	\centering
	\includegraphics[scale=0.88]{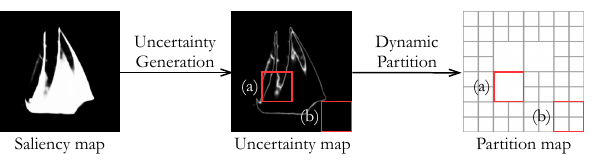 }
	\vspace{-2em}\caption{The illustration of proposed Adaptive Dynamic Partition (ADP) mechanism. 
		(a) In the blur region, we discontinue further partition to ensure the efficacy of uncertainty guidance; 
		(b) For the clear region, we perform further partition to minimize computational expenses. 
	}
	\label{fig:adp_demonstration}
\end{figure}

Third, we design an Adaptive Dynamic Partition (ADP) mechanism, 
that adapts the uncertainty guidance to the case of a large spatial scale. 
Due to the small proportion of uncertainty regions in the saliency map (around 2\% - 5\%), 
calculating the mask attention with uncertainty guidance globally would incur prohibitively high computational costs, 
which is obviously unfavorable for the refinement of uncertainty guidance at larger spatial scales. 
Inspired by recent research on dynamic inference for data diversity \cite{huang2022glance,han2021dynamic}, 
we introduce the dynamic inference for the SOD task. 
Specifically, we apply distinct processes based on the varying degrees of uncertainty within different regions in the saliency map. 
As shown in Fig. \ref{fig:adp_demonstration}, in the case of sharper regions, 
we perform further partition to minimize computational expenses. 
Correspondingly, for regions with higher blur, we discontinue further partition to ensure the efficiency of uncertainty guidance. 
With the aid of the ADP mechanism, we harness the full potential of uncertainty guidance, 
thereby further improving the model performance in fine-grained saliency prediction. 

Our main contributions can be summarized as follows:

\begin{itemize}
	\item 
	We design two simple yet effective modules, Multilevel Interaction Attention (MIA) and 
	Scale Spatial Consistent Attention (SSCA), effectively utilizing multilevel 
	features to accurately localize salient objects in the scene. 
	\item To alleviate the limitation of boundary guidance,  we instead introduce an uncertainty guidance learning approach to SOD, 
	and design the uncertainty refinement attention (URA) module 
	to boost the model's perception capability of uncertainty areas. 
	\item We propose an Adaptive Dynamic Partition (ADP) mechanism to reduce computational cost 
	and adapt the uncertainty refinement approach to large spatial scales.

\end{itemize}

To the best of our knowledge, we proposed for the first time a solution to the long-overlooked problem 
of widespread unsaturated areas and artifacts in saliency object prediction via the utilization of uncertainty-guided dynamically iterative refinement. Through the integration of these suggested modules, 
we have developed a new network specifically designed for fine-grained saliency object detection, 
and demonstrated its effectiveness quantitatively and qualitatively on seven challenging datasets. 

\section{Related Work}

Over the past few decades, numerous methods for salient object detection (SOD) 
have been proposed and have demonstrated promising results on various benchmark datasets. 
These existing SOD methods can be roughly categorized into feature enhancement and aggregation methods,
boundary guided methods, and uncertainty guided methods. 
\subsection{Feature enhancement and aggregation methods for SOD}
As SOD represents a dense detection task, numerous methods have been developed to address the challenge of compensating for the absence of local details in high-level features by integrating low-level features, 
thereby obtaining high-quality saliency maps. 
PoolNet\cite{PoolNet} adopted the Feature Pyramid Network (FPN) structure and delivered high-level semantic information to low-level feature maps, 
remedying the drawback of U-shape\cite{Unet} structure that top-down signals are gradually diluted. 
MiNet \cite{MiNet} let adjacent level features interact with each other, 
aiming to minimize the discrepancies and improve the spatial coherence of multilevel features. 
ICON\cite{ICON} incorporated convolution kernels with different shapes to enhance the diversity of multilevel features. 
BBRF\cite{BBRF} designed a switch-path decoder with various receptive fields to deal with large or small-scale objects. 
\begin{figure*}
	\centering
	\includegraphics[scale=0.88]{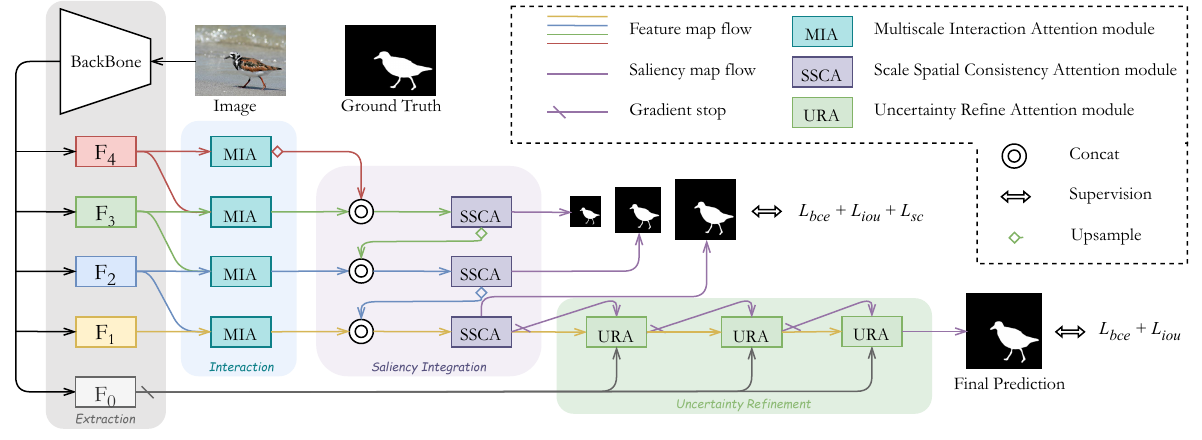}
	\vspace{-1em}\caption{Architecture of the proposed model. The backbone is defined as a hierarchical network structure (e.g. ResNet\cite{ResNet}, SwinTransformer\cite{Swin}). The multilevel features extracted by the backbone are denoted as F$_0$ - F$_4$ and the spatial dimensions decrease sequentially. }
	\label{fig:overview}
\end{figure*}

Although there have been extensive investigations into feature enhancement 
and aggregation in various methods, a majority of them primarily concentrate on 
the aggregation of features, neglecting the crucial integration of 
salient information across multiple scales within the aggregated features. 
{We posit that this deficiency results in inaccurate localization of salient objects, contributing to the emergence of shadows and unsaturated regions in saliency prediction. To overcome this, 
we first design the Multilevel Interaction Attention module to facilitate interaction and perception among features at different levels, intending to leverage better global saliency representation in high-level features to reduce non-salient noise within low-level features. }
Then, we introduce the Scale Spatial Consistency Attention module to 
integrate salient information at different scales within the aggregated features, 
obtaining consistent and integral salient information across different spatial scales. 
\subsection{Boundary guiding methods for SOD}
Boundary guidance is widely acknowledged as a crucial cue for improving the outcomes of SOD 
due to its ability to not only assist in the accurate localization of salient objects but also refine local details. 
For example, VST\cite{VST} designed a multi-task decoder, which utilizes boundary information to guide the localization of salient objects. 
Amulet\cite{Amulet} provided boundary refinements by adding short connections from the boundary information to the predicted results. 

It needs to be acknowledged that boundary guidance effectively enhances 
the performance of the models proposed by previous SOD methods. 
However, boundary guidance has the following inherent flaws: 
\begin{itemize}
\item Boundary information learned by the model not only provides boundary information of salient objects, 
but also brings redundant interference from non-salient objects, particularly in complex environments. 
This may potentially mislead the model. 
\item Boundary guidance based on prior knowledge provided during all stages of training and inference remains fixed, 
which cannot be adapted to the model prediction of where the actual unsaturated region is located. 
\item With the powerful global modeling capability of the Transformer, 
the previous challenges in SOD tasks, such as object localization, have been well addressed. 
The assistance that boundary information can provide in auxiliary positioning or refining details becomes very limited. 
\end{itemize}
\subsection{Uncertainty guiding methods for SOD}
Uncertainty guiding for SOD has not been sufficiently explored. Among the limited existing methods, 
UCF\cite{Zhang_2017_ICCV} learn deep uncertain convolutional features to encourage the robustness and accuracy of saliency detection. 
ISPRN\cite{kim2022revisiting} first introduced the method of generating uncertainty maps to SOD, 
and blended them with foreground and background as contextual information to provide guidance for the model. 
RCSBN\cite{Ke_2022_WACV} assigned larger weights to uncertain predicted pixels in the loss functions, 
which can also be considered as an uncertainty guiding based approach. 

Unlike previous methods, we employ a more explicit approach in leveraging uncertainty information to guide the model learning. 
Specifically, we begin by generating an uncertainty map using the predicted saliency map. 
Then, we utilize this uncertainty map to mask the attention, enabling the model to focus solely on the uncertainty regions. 
As a result, the model's perception of uncertain areas is strengthened, leading to enhanced predictive accuracy. Besides, to strike a balance between the guiding effect and computational cost, 
we introduce an adaptive dynamic partition mechanism, which can partition the images based on the current prediction result, 
thereby substantially improving the model prediction performance. 
\section{Methodology}
\subsection{Overview}
\label{Overview}

\begin{figure*}
	\centering
	\includegraphics[scale=0.88]{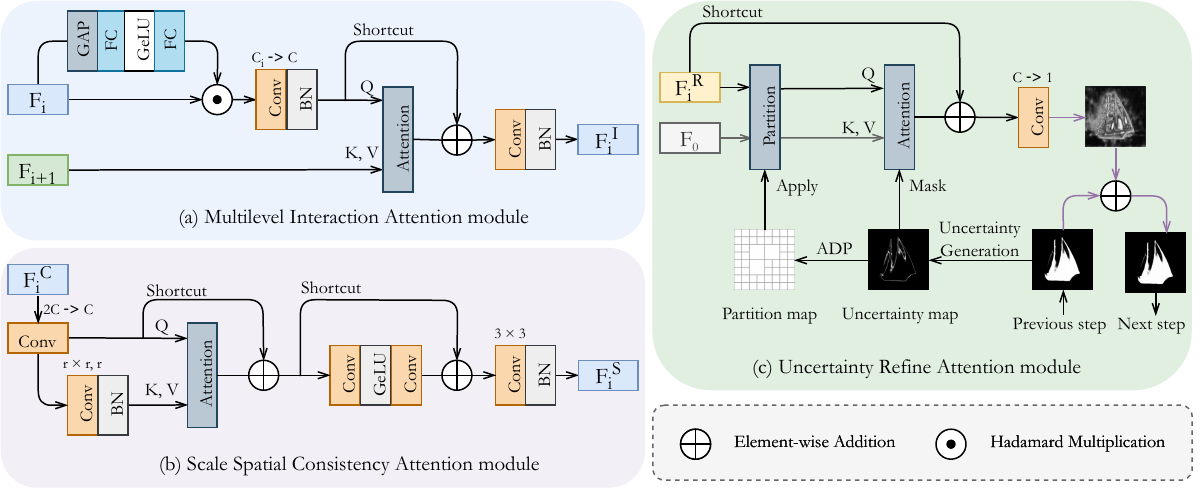}
	\caption{Details of proposed modules. All convolutions, excluding the labeled ones, 
		utilize a 1$\times$1 kernel size, a stride of 1, and preserve the channel dimension. }

	\label{fig:modules}
\end{figure*}
Fig. \ref{fig:overview} provides an overview of our model. 
As can be seen, for each input image, the model will process it through four stages: Extraction, Interaction, Saliency Integration, and Uncertainty Refinement. 
We first use a hierarchical backbone to extract multilevel features, 
and then pass the features through the Multilevel Interaction Attention (MIA) module for interaction and perception. 
Afterward, we use the proposed Scale Spatial Consistent Attention (SSCA) module to integrate saliency information across different levels from aggregated features. 
Finally, in order to reduce artifacts in the predicted image and further improve its reliability, 
we utilize the proposed Uncertainty Refinement Attention (URA) module for uncertainty guided refinement. 

We have extensively employed attention mechanisms in the proposed modules, defined as:
\begin{align}
	\small
	\begin{split}
		&{Attention}(Q,K,V) = Softmax(Q\otimes K^T)\otimes V,\\
	\end{split}
\end{align}
where $\otimes$ denotes matrix multiplication, $Q$, $K$, and $V$ denotes query, key, and value defined in \cite{Attentionisallyouneed}. 
\subsection{Multilevel Interaction Attention}
\label{text:MIA}
Recent works \cite{ICON,kim2022revisiting,BBRF} have shown that employing multiple receptive fields can help the network learn features that capture objects with various sizes. 
However, these methods apply the same processing to features of different scales and ignore their distinctive properties. 
{Unlike them, we attempt to facilitate the use of complementarity between features of different scales by employing interactive attention, 
thereby replacing traditional feature enhancement methods. }
In comparison to MiNet\cite{MiNet}, the proposed MIA avoids the need for 
upsampling or downsampling features from different levels, 
thereby mitigating the risk of information loss. 
Besides, we conducted comprehensive experiments to explore various interactive connections. Detailed discussions can be found in the experimental section. 

Channel attention mechanisms\cite{hu2018squeeze,yang2020gated} have demonstrated promising results in image classification. 
Some SOD methods, such as ICON\cite{ICON} and PFSNet\cite{ma2021pyramidal}, incorporate channel attention mechanisms into their networks, 
resulting in improved network performance. 
As shown in Fig. \ref{fig:modules} (a), we first leverage recent advancements to optimize the features extracted by the backbone using channel attention mechanisms. 
Then, we introduce higher-level feature information to enable effective interaction and perception among different hierarchical features by using attention mechanisms. 
The whole procedure is summarized as follows: 
\begin{align}
	\small
	\begin{split}
		&\hat{F_i} = \chi(F_i \odot \psi(GeLU(\psi(GAP(F_i)))),\\
		&Q = \psi(\hat{F_i}), K = \psi(F_{i+1}), V = \psi(F_{i+1}),	\\
		&F_i^I = BN(\chi(\hat{F_i}+Attention(Q,K,V))).
	\end{split}
\end{align}
Here, $\chi$, $\psi$, and $BN$ represent 1$\times$1 convolutions, linear projections, and BatchNorm, respectively. 
$\hat{F_i}$ denotes $F_i$ after applying channel attention, and a 1$\times$1 convolution is employed to convert its channel count to C (set to 64 as default). 
{Additionally, since $F_4$ is the highest-level feature without any further higher-level features to interact with, we exclusively apply channel attention to $F_4$. }

{The proposed MIA module distinguishes itself from existing feature enhancement methods (FEMs)\cite{ICON,ASPP,zhao2017pspnet,Liu_2018_ECCV} in several key ways: 1. Unlike previous convolutional methods, the MIA module is not constrained by the convolutional local receptive field; it can globally enhance features, accentuating salient information within low-level features. 2. Existing FEMs often treat features of different scales equally, whereas {MIA fully considers the distinction of features at different levels} and applies distinctive treatments to them, resulting in complementary advantages among the features. The experimental section validates the superiority of MIA over existing FEMs. }

\subsection{Scale Spatial-Consistent Attention}
Following processing by MIA, we aggregate features from different levels and reduce their channel dimension to C through a convolution. The result is defined as F$_i^C$. 
In order to further integrate salient information from the aggregated features, VST\cite{VST} introduced the self-attention mechanism into the decoder. 
Self-attention mechanism can effectively integrate salient information, but it also has two drawbacks: 
1. It neglects the inconsistency of salient information within aggregated features from different spatial scales; 
2. It incurs massive computational costs. 

Previous SOD methods\cite{xu2021locate,MiNet} commonly utilize multiscale downsampling\cite{zhao2017pspnet} 
to process features while integrating salient information, 
in order to extract more comprehensive information. 
Assume that low-resolution feature maps provide a better representation of global information, 
we uniformly downsample the aggregated features to a same small resolution, 
aiming to achieve a global-level saliency representation. 
Specifically, as shown in Fig. \ref{fig:modules} (b), we adopt a convolution with an $r \times r$ kernel 
and a stride of $r$ to decrease the spatial scale of the features when generating $K$ and $V$ for attention calculation. 
In this case, $r$ is assigned a value of $2^{3-i}$ for F$_i^C$, where $1 \le i \le 3$. This indicates that the feature, 
following the spatial reduction, will possess the same resolution as F$_3$. 
Then, we promote the integration of salient information through attention mechanisms. 
The whole procedure can be described as follows: 
\begin{align}
	\small
	\begin{split}
		&\hat{F_i}^C = BN(\chi_{rxr,r}(F_i^C)),\\
		&Q = \psi(F_i^C), K = \psi(\hat{F_i}^C), V = \psi(\hat{F_i}^C),	\\
		&\hat{F_i}^S = F_i^C + Attention(Q,K,V),\\
		&F_i^S = BN(\chi(\hat{F_i}^S + \chi(GeLU(\chi(\hat{F_i}^S))))).
	\end{split}
\end{align}
This approach enhances the consistency of salient information with a lower computational cost. 
Furthermore, at the end of each SSCA module, we acquire the saliency prediction through a 1$\times$1 convolution for the purpose of multilevel supervision. 

The proposed SSCA module distinguishes itself from existing feature aggregation methods\cite{VST,MiNet,ma2021pyramidal} in several major aspects: 
1. 
{Existing methods focus primarily on designing various paths for feature aggregation, while neglecting the crucial integration of salient information within the aggregated features. 
In contrast, SSCA achieves this by downsampling features to a small resolution, 
extracts and integrates features that exhibit saliency across different spatial scales, thereby ensuring the consistency and integrity of the salient objects. 
2. While existing methods simultaneously promote feature aggregation and local detail recovery, our model delegates the task of local detail recovery to uncertainty refinement. }Therefore, this allows SSCA to focus more accurately on localizing salient objects. The experimental section confirms the superiority of SSCA over existing feature aggregation methods.

\subsection{Uncertainty Refinement Attention}
\begin{figure}
	\centering
	\includegraphics[scale=0.88]{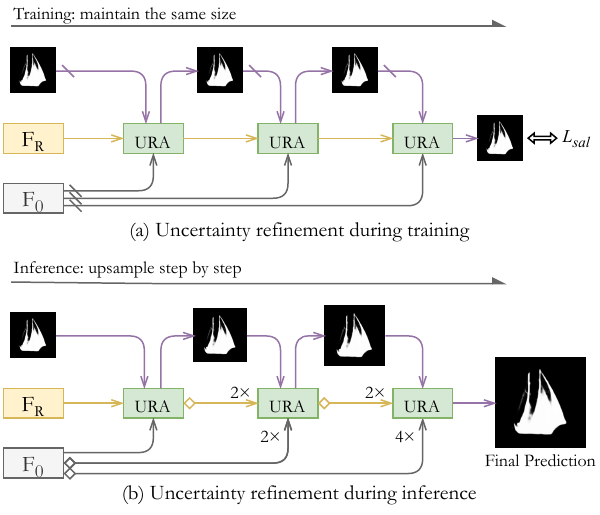}
	\vspace{-2em}\caption{The feature maps are progressively upsampled during the inference process until they align with the original image size. 
	The upsample ratio is variable and can be adjusted based on specific requirements. 
}
	\label{fig:ur}
\end{figure}
The Uncertainty Refinement Attention (URA) module aims to mitigate artifacts and undersaturation regions in saliency prediction by leveraging uncertainty guidance and low-level features. 
To achieve this, we first generate an uncertainty map from the previous stage prediction, formulated as: 
\begin{equation}
	\small
	\begin{split}
		&	\hat{U} = t-abs(S-t),\\
		&	U = \chi^G(\hat{U}). 
	\end{split}
	\label{eq:ug}
\end{equation}
Here, $S$ denotes the previous stage saliency prediction, $U$ denotes the uncertainty map required, {$abs$ indicates taking the absolute value, }
$t$ denotes the threshold and we set it as a constant value of 0.5. 
For the prediction of a point in $S$, it is considered certain when its value is close to 0 or 1.
Conversely, if the value is close to 0.5, it is considered uncertain. 
We apply Gaussian convolution $\chi^G$ with parameters $k=7$ and $\sigma=1$ for smoothing. 
The process defined in \eqref{eq:ug} is referred to as Uncertainty Generation, which can reveal the uncertain prediction regions in $S$. 
Besides, we use the saliency prediction from the last SSCA module in the first URA module. 

Then, we utilize the uncertainty map to mask the attention. The mask matrix and mask attention are calculated as: 
\begin{align}
	\small
	\begin{split}
		&M(x,y) = \left \{ 
		\begin{matrix} 
			&	0 & if\ U(x,y)>0.01, \\ 
			&	-\infty & otherwise, 
		\end{matrix}\right.\\
		&MaskAttention(M,Q,K,V) = Softmax(M+Q\otimes K^T)\otimes V,
	\end{split}
\end{align}

Thus the attention is calculated as:
\begin{align}
\label{eq:mask_attention}
	\small
	\begin{split}
		&Q = \psi(F_i^R), K = \psi(F_0), V = \psi(F_0),	\\
		&F_{i+1}^R = BN(\chi({F_i}^R+MaskAttention(M,Q,K,V))). 
	\end{split}
\end{align}
Here, $F_i^R$ denotes the feature for refinement, and we use $F_1^S$ as $F_0^R$ initially. 
F$_0$ serves as a low-level feature, extracted as the initial feature by the backbone. 
It encompasses abundant local semantic information, 
making it a suitable reference for uncertainty refinement. 
At last, $F_{i+1}^R$ is transformed into a one-dimensional representation using a 1$\times$1 convolution. 
This transformed representation is then added to the previous prediction, 
resulting in the filling of unsaturated regions and the elimination of shadows within the prediction. 

{
It is important to note that mask operations are a common technique in image processing, and mask attention calculation was previously proposed in Mask2Former~\cite{cheng2022masked}. However, we use mask attention for a different purpose. }
Specifically, Mask2Former aims to predict masks for different object categories, serving more as a feature extraction tool. In contrast, our approach refines the model's attention by using uncertainty masks, which restrict attention to uncertain regions, providing a more explicit refinement method. In other words, our method can be viewed as a deep-learning-based post-processing technique. Furthermore, we have designed an Adaptive Dynamic Partitioning mechanism (ADP), which extends the mask attention calculation to larger spatial scales.
Fig. 4 (c) provides an illustration of the URA module. We will introduce ADP in the next subsection. 
So for now, we can temporarily disregard the presence of ADP and partition in the figure. 

We aim to design URA in such a way that it can effectively 
address the artifacts and undersaturated regions 
in the predicted image by leveraging low-level features. 
The uncertainty map mask plays a crucial role in achieving this objective. 
In contrast to previous methods{\cite{{Ke_2022_WACV},F3Net}} that weight complex regions in the loss function, 
our URA offers a more explicit learning approach, 
enabling the model to focus on suppressing the uncertain regions that occurred in previous stage predictions 
and thus improve the quality of the predicted image iteratively. 
The whole Uncertainty Refinement includes three consecutive uncertainty refinement attention modules. 
As depicted in Fig \ref{fig:ur}, we keep the size of the feature maps unchanged 
during training to expedite the training process. 
{However, to improve the utilization of uncertainty guidance and enhance the fineness of the prediction map, we progressively upsample the feature maps during inference until they align with the original image size. The cornerstone of this design is the spatial scale insensitivity of the attention mechanism, which necessitates only ensuring the uniformity of input features across channel dimensions. We will further discuss the inconsistency between training and inference stemming from this design in the next section. }

\subsection{Adaptive Dynamic Partition}
We found through our experiments that, on average, only 2\%-5\% of the image predictions fall within the uncertainty regions. Therefore, conducting attention modeling on the whole image will 
lead to considerable computational inefficiency and hinder processing at higher resolutions. 
To overcome this, we design an Adaptive Dynamic Partition (ADP) mechanism, 
which can dynamically partition the images based on the current prediction, 
significantly reduce computational costs while ensuring the effectiveness of uncertainty guidance. 

Specifically, for a saliency map, the level of uncertainty differs across various regions. 
As shown in Fig. \ref{fig:adp_demonstration}, in the case of sharper regions, 
further partition can be applied to minimize computational expenses. 
However, for regions with higher blur, we perform attention calculation without additional partition, 
which may fragment uncertain information and impede the model's perception of uncertain regions. 

\begin{algorithm}
\caption{{Adaptive Dynamic Partition}}
\label{alg:ADP}
\begin{algorithmic}[1]
\State \textbf{Note:}
\State {Function \textit{Partition}: shape conversion for features}
\State {Function \textit{Att}: mask attention for features}
\State \textbf{Input:} $x$ \Comment{Features, size=$(B,C,H,W)$}
\State \textbf{Input:} $l$ \Comment{Low-level features, size=$(B,C,H,W)$}
\State \textbf{Input:} $u$ \Comment{Uncertainty map, size=$(B,1,H,W)$}
\State \textbf{Parameter:} $pthreshold$ \Comment{Threshold for partition}
\State \textbf{Parameter:} $minsize$ \Comment{Minimum size for partition}
\State \textbf{Output:} $x$
\Function {ADP}{$x$, $l$, $u$}
\State $(h, w) \gets $\textit{shape}$(x)$[2:]
\State $x_{w} \gets $\textit{Partition}$(x)$
\State $l_{w} \gets $\textit{Partition}$(l)$
\State $u_{w} \gets $\textit{Partition}$(u)$
\For{$i$ from $0$ to $4$}
\State $p \gets \frac{\textit{Sum}(u_{w}[i])}{(h*w)}$
\If{$(p < pthreshold$ and $h > minsize)$}
\State $x_{w}[i] \gets $\textit{ADP}$(x_{w}[i],l_{w}[i],u_{w}[i])$
\Else
\State $x_{w}[i] \gets $\textit{Att}$(x_w[i],l_w[i],u_w[i])$
\EndIf
\EndFor
\State $x \gets $\textit{reverse}$(x_{w})$
\Return $x$
\EndFunction
\end{algorithmic}
\end{algorithm}

{Algorithm \ref{alg:ADP} offers an illustration of the proposed ADP.} To provide a more accessible description, 
the proposed ADP can be conceptualized as a combination of global attention and window attention\cite{Swin}. 
The key difference is that window attention conducts window partition before attention calculation, 
whereas ADP dynamically determines the need for further partition based on the uncertainty level of the current window. 
Technically, we measure the level of uncertainty in a window by calculating the proportion $p$ that 
represents the percentage of uncertainty regions within the window. 
We also define $p_t$ as the threshold for further partition of windows and set it to 0.2 (20\%). 
When $p<p_t$, windows undergo further partition; Otherwise, no further partition takes place. 
In addition, setting $p_t$ to 0 means no partition will be performed, 
and the attention calculation will be performed globally. 
Conversely, setting $p_t$ to 1 will result in window attention calculation, 
where the feature map will be divided into equally sized windows. 
We can adjust $p_t$ according to various requirements, 
and the minimum partition size is set to 1/32 of the input image. 

Furthermore, as previously mentioned, the URA module demonstrates 
dynamic reasoning characteristics under the support of the ADP mechanism. 
Fig. \ref{fig:modules} (c) provides a simplified illustration, 
where partition and attention calculation occur only once. 
However, during actual inference, the process of partition and attention 
calculation is recursively performed multiple times until all windows 
become relatively blurry or reach the minimum partition size. 
{Moreover, this dynamic reasoning capability empowers the model to meet features with different spatial scales during training, enabling it to adjust to the spatial scale diversity of features and thus resolve the inconsistency between training and inference mentioned in the previous section. }
\begin{table*}
	\scriptsize
	\centering
	\setlength\tabcolsep{0.05mm}
	\caption{Quantitative comparison of our proposed model with other SOTA SOD methods on six benchmark datasets. The best results are shown in \textbf{bold}. The symbols “↑”/“↓” mean that a higher/lower score is better. `-R', '-R2', and `-S' means the ResNet50, Res2Net50\cite{gao2019res2net}, and SwinTransformer backbone. }
\begin{tabular}{l|c|c|cccc|cccc|cccc|cccc|cccc|cccc}%
	\hline 
	\multicolumn{3}{c|}{dataset} & \multicolumn{4}{c|}{DUT-O} & \multicolumn{4}{c|}{DUTS} & \multicolumn{4}{c|}{ECSSD} & \multicolumn{4}{c|}{HKU-IS} & \multicolumn{4}{c|}{PASCAL-S} & \multicolumn{4}{c}{SOD}\\
	\hline 
	Method & {MACs} & {Params} & $M\downarrow$ & $E_\xi^{m}\uparrow$ & $S_m\uparrow$ & $F_\beta^w\uparrow$ & $M\downarrow$ & $E_\xi^{m}\uparrow$ & $S_m\uparrow$ & $F_\beta^w\uparrow$ & $M\downarrow$ & $E_\xi^{m}\uparrow$ & $S_m\uparrow$ & $F_\beta^w\uparrow$ & $M\downarrow$ & $E_\xi^{m}\uparrow$ & $S_m\uparrow$ & $F_\beta^w\uparrow$ & $M\downarrow$ & $E_\xi^{m}\uparrow$ & $S_m\uparrow$ & $F_\beta^w\uparrow$ & $M\downarrow$ & $E_\xi^{m}\uparrow$ & $S_m\uparrow$ & $F_\beta^w\uparrow$ \\
	\hline 
	\multicolumn{25}{c}{CNN Methods}\\
	\hline 
	CPD-R & {17.77} & {47.85} & .056 & .868 & .825 & .719 & .043 & .914 & .869 & .795 & .037 & .951 & .918 & .898 & .034 & .95 & .905 & .875 & .071 & .891 & .848 & .794 & .11 & .849 & .771 & .713 \\
	PoolNet & {88.89} & {68.26} & .054 & .867 & .831 & .725 & .037 & .926 & .887 & .817 & .035 & .956 & .926 & .904 & .03 & .958 & .919 & .888 & .065 & .907 & .865 & .809 & .103 & .867 & .792 & .746 \\
	CSF-R2 & {18.96} & {36.53} & {.055} & .869 & .838 & .733 & .037 & .93 & .89 & .823 & .033 & .96 & .93 & .91 & .03 & .96 & .921 & .891 & .069 & .899 & .862 & .807 & .098 & .872 & .8 & .757 \\
	ITSD-R & {15.96} & {26.47} & .061 & .88 & .84 & .75 & .041 & .929 & .885 & .824 & .034 & .959 & .925 & .91 & .031 & .96 & .917 & .894 & .066 & .908 & .859 & .812 & .093 & .873 & .809 & .777 \\
	MINet-R & {87.11} & {126.38} & .056 & .869 & .833 & .738 & .037 & .927 & .884 & .825 & .033 & .957 & .925 & .911 & .029 & .96 & .919 & .897 & .064 & .903 & .856 & .809 & .092 & .87 & .805 & .768 \\
	LDF & {15.51} & {25.15} & \textbf{.052} & .869 & .839 & .752 & {.034} & .93 & .892 & {.845} & .034 & .954 & .924 & .915 & .028 & .958 & .919 & .904 & .06 & .908 & .863 & .822 & .093 & .866 & .8 & .765 \\
	
	PFSNet & {45.41} & {31.18} & .055 & .878 & .842 & .756 & .036 & .931 & .892 & .842 & .031 & .958 & .93 & .92 & {.026} & .962 & .924 & .91 & .063 & .906 & .86 & .819 & .089 & .875 & .81 & .781 \\
	ICON-R & {20.91} & {33.09} & .057 & \textbf{.884} & {.844} & {.761} & .037 & {.932} & .889 & .837 & .032 & .96 & .929 & .918 & .029 & .96 & .92 & .902 & .064 & .908 & .861 & .818 & \textbf{.084} & \textbf{.882} & \textbf{.824} & \textbf{.794} \\
	
	\rowcolor{gray!20} Ours-R & {24.18} & {37.38} & .061 & .877 & \textbf{.846} & \textbf{.766} & \textbf{.033} & \textbf{.942} & \textbf{.906} & \textbf{.865} & \textbf{.029} & \textbf{.961} & \textbf{.934} & \textbf{.922} & \textbf{.025} & \textbf{.966} & \textbf{.932} & \textbf{.917} & \textbf{.055} & \textbf{.92} & \textbf{.876} & \textbf{.84} & \textbf{.084} & .864 & .814 & .776 \\ 
	
	\hline 
	\multicolumn{25}{c}{Transformer Methods}\\
	\hline 
	
	VST & {23.16} & {44.63} & .058 & .89 & .852 & .758 & .037 & .94 & .897 & .831 & .032 & .965 & .934 & .911 & .029 & .968 & .928 & .897 & .06 & .919 & .874 & .821 & .085 & .876 & .82 & .776 \\
	ICON-S & {52.59} & {94.30} & \textbf{.043} & {.906} & {.869} & .804 & {.025} & .96 & .917 & .886 & .023 & .972 & .941 & .936 & .022 & .974 & .935 & .925 & .048 & .93 & {.885} & .854 & .083 & {.885} & .825 & .802 \\ 
	ISPRNet-S & {101.43} & {90.72} & .046 & {.903} & {.875} & {.799} & {.024} & {.963} & \textbf{.931} & {.891} & {.023} & {.972} & {.949} & {.937} & {.021} & {.977} & {.944} & {.927} & {.048} & {.932} & \textbf{.893} & {.857} & {.082} & {.881} & {.834} & {.802} \\
	BBRF & {46.00} & {74.40} & .044 & {.899} & {.861} & {.803} & {.025} & {.952} & {.909} & {.886} & {.022} & {.972} & {.939} & {.944} & {.02} & {.972} & {.932} & {.932} & {.049} & {.927} & {.878} & {.856} & .078 & .868 & .822 & .802\\
	SRformer & {12.82} & {90.70} & \textbf{.043} & {.894} & {.86} & .784 & {.027} & {.952} & .91 & .872 & .028 & .961 & .932 & .922 & .025 & .966 & .928 & .912 & {.051} & {.924} & {.878} & {.845} & .088 & .862 & .809 & .77 \\ 
	\rowcolor{gray!20} Ours-S & {61.76} & {96.54} & .045 & \textbf{.911} & \textbf{.878} & \textbf{.818} & \textbf{.022} & \textbf{.964} & \textbf{.931} & \textbf{.906} & \textbf{.02} & \textbf{.975} & \textbf{.95} & \textbf{.946} & \textbf{.019} & \textbf{.978} & \textbf{.945} & \textbf{.937} & \textbf{.046} & \textbf{.934} & {.892} & \textbf{.863} & \textbf{.075} & \textbf{.89} & \textbf{.838} & \textbf{.818} \\
	\hline 
\end{tabular}
	\label{tab:quantitative_comparison}
\end{table*}

\begin{figure*}
	\centering
	\includegraphics[scale=0.24]{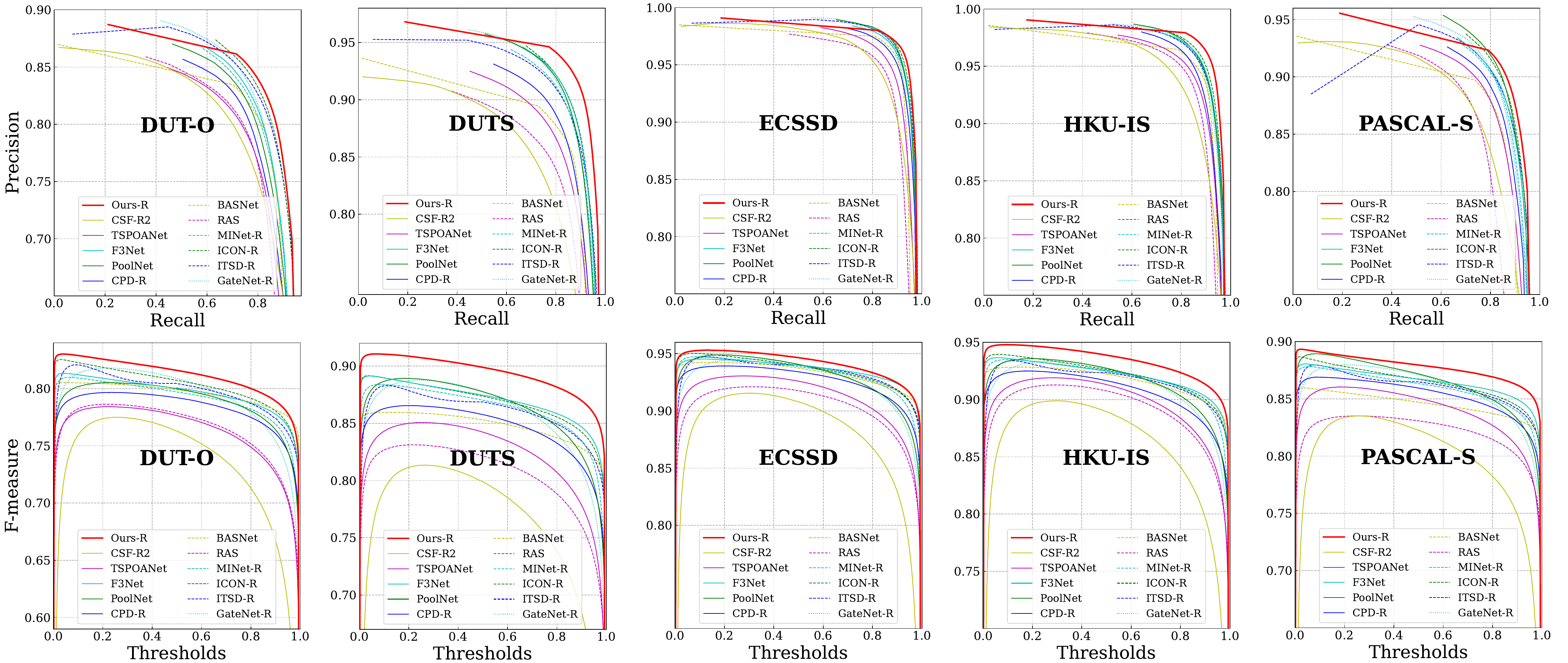}
	\vspace{-0.5em}\caption{Precision-Recall curves (row 1) and F-measure curves (row 2) of Ours-R and other CNN SOTA methods on five benchmark datasets. 
	}	

	\label{fig:pr&fm}
\end{figure*}
\subsection{Supervision Strategy}
We train our network using Binary Cross Entropy (BCE)\cite{de2005tutorial} loss and Intersection over Union (IoU) loss,
which has been widely adopted in related methods such as \cite{VST,ICON}. 
BCE loss calculates the loss independently for each pixel, which is defined as:
\begin{align}
	\small
	\begin{split}
		\mathcal{L}_{bce}=&-\sum_{x=1}^{H}\sum_{y=1}^{W}[G(x,y)logP(x,y)\\
		&+(1-G(x,y))log(1-P(x,y))],\\
	\end{split}
\end{align}
where $H$, $W$ are the height and width of the image, and $P(x,y)$, $G(x,y)$ denote 
the pixels of the prediction and the ground truth at position $(x,y)$, respectively. 
Meanwhile, the IoU loss considers the inter-pixel relationship and 
effectively addresses the limitations of BCE, and can be formulated as: 
\begin{align}
	\small
	\begin{split}
		\mathcal{L}_{iou}=&1-\frac{\sum_{x=1}^{H}\sum_{y=1}^{W}(P(x,y)G(x,y))}{\sum_{x=1}^{H}\sum_{y=1}^{W}(P(x,y)+G(x,y)-P(x,y)G(x,y))}.
	\end{split}
\end{align}
Besides, we reinforce the consistency between saliency predictions generated by different
SSCA modules via the spatial consistency loss function, which is defined as: 
\begin{align}
	\small
	\begin{split}
		\mathcal{L}_{sc}=&\sum_{x=1}^{H}\sum_{y=1}^{W}\left|\left| S_i(x,y) - \hat{S}_{i+1}(x,y) \right|\right|,
	\end{split}
\end{align}
where $S$ denotes saliency prediction from the proposed SSCA, and $\hat{S}$ 
represents saliency prediction once gradient stopping has been applied. 
We employ spatial consistency loss aims to ensure consistency between 
low-level and high-level outputs, while preserving the integrity 
of high-level outputs from low-level gradient flow effects. 

We refrained from incorporating any form of weighting in the loss function. 
Interestingly, we discovered that the inclusion of weighted loss could 
impede the training process and impose limitations on the model performance, 
especially when the network performance is already satisfactory. 
A more detailed discussion can be found in the experimental section. 
During training, we specifically use the multilevel supervision strategy to facilitate the training process. 
The total loss function is defined as:
\begin{align}
	\small
	\begin{split}
		\mathcal{L}_{tot}=&\sum_{i=1}^{3}(\mathcal{L}_{bce}(S_i,G)+\mathcal{L}_{iou}(S_i,G))\\
		+ &\sum_{i=1}^{3}(\mathcal{L}_{bce}(R_i,G)+\mathcal{L}_{iou}(R_i,G))\\
		+ &\sum_{i=1}^{2}\mathcal{L}_{sc}(S_i,\hat{S}_{i+1}), 
	\end{split}
\end{align}
$R$ denotes refined prediction from the proposed URA. 
Prior to calculating the loss, 
$S$ and $R$ are initially upsampled to match the shape of the ground truth $G$ via bilinear interpolation. 

\section{Experiment}

\subsection{Implementation Details}
\begin{table*}
	\scriptsize
	\centering
	\setlength\tabcolsep{1mm}
	\caption{Comparison of our proposed model with other SOTA CNN-based SOD methods on SOC test set. The best results are shown in \textbf{bold}. }
\begin{tabular}{l|r|ccccccccccccccccc|c}%
	\hline 
	{Attr} & Metrics & NLDF & C2SNet & SRM & R3Net & BMPM & DGRL & PiCA-R & RANet & AFNet & CPD & PoolNet & EGNet & BANet & SCRN & PFSNet & ICON-R & MENet & Ours-R \\
	\hline 
	\multirow{4}{*}{AC} & $M\downarrow\:$ & .119 & .109 & .096 & .135 & .098 & .081 & .093 & .132 & .084 & .083 & .094 & .085 & .086 & .078 & .073 & \textbf{.062} & .069 & \textbf{.062} \\
						& $E_\xi^{m}\uparrow\:$ & .784 & .807 & .824 & .753 & .815 & .853 & .815 & .765 & .852 & .843 & .846 & .854 & .858 & .849 & .876 & \textbf{.891} & .86 & \textbf{.891} \\
						& $S_m\uparrow\:$ & .737 & .755 & .791 & .713 & .780 & .790 & .792 & .708 & .796 & .799 & .795 & .806 & .806 & .809 & .821 & .835 & .812 & \textbf{.836} \\
						& $F_\beta^w\uparrow\:$ & .620 & .647 & .690 & .593 & .680 & .718 & .682 & .603 & .712 & .727 & .713 & .731 & .740 & .724 & .768 & \textbf{.784} & .746 & .772 \\
	\hline 
	\multirow{4}{*}{BO} & $M\downarrow\:$ & .354 & .267 & .306 & .445 & .303 & .215 & .200 & .454 & .245 & .257 & .353 & .373 & .271 & .224 & .21 & .200 & .255 & \textbf{.144} \\
						& $E_\xi^{m}\uparrow\:$ & .539 & .661 & .616 & .419 & .620 & .725 & .741 & .404 & .698 & .665 & .554 & .528 & .650 & .706 & .811 & .740 & .765 & \textbf{.822} \\
						& $S_m\uparrow\:$ & .568 & .654 & .614 & .437 & .604 & .684 & .729 & .421 & .658 & .647 & .561 & .528 & .645 & .698 & .712 & .714 & .659 & \textbf{.779} \\
						& $F_\beta^w\uparrow\:$ & .622 & .730 & .667 & .456 & .670 & .786 & .799 & .453 & .741 & .739 & .610 & .585 & .720 & .778 & .792 & .794 & .731 & \textbf{.851} \\
	\hline 
	\multirow{4}{*}{CL} & $M\downarrow\:$ & .159 & .144 & .134 & .182 & .123 & .119 & .123 & .188 & .119 & .114 & .134 & .139 & .117 & .113 & \textbf{.105} & .113 & .113 & {.109} \\
						& $E_\xi^{m}\uparrow\:$ & .764 & .789 & .793 & .710 & .801 & .824 & .794 & .715 & .802 & .821 & .801 & .790 & .824 & .820 & \textbf{.841} & .829 & .816 & {.837} \\
						& $S_m\uparrow\:$ & .713 & .742 & .759 & .659 & .761 & .770 & .787 & .624 & .768 & .773 & .760 & .757 & .784 & .795 & .796 & .789 & .778 & \textbf{.798} \\
						& $F_\beta^w\uparrow\:$ & .614 & .655 & .665 & .546 & .678 & .714 & .692 & .542 & .696 & .724 & .681 & .677 & .726 & .717 & \textbf{.737} & .732 & .717 & \textbf{.737} \\
	\hline 
	\multirow{4}{*}{HO} & $M\downarrow\:$ & .126 & .123 & .115 & .136 & .116 & .104 & .108 & .143 & .103 & .097 & .100 & .106 & .094 & .096 & .088 & .092 & \textbf{.084} & \textbf{.084} \\
						& $E_\xi^{m}\uparrow\:$ & .798 & .805 & .819 & .782 & .813 & .833 & .819 & .777 & .834 & .845 & .846 & .829 & .850 & .842 & .862 & .852 & .873 & \textbf{.875} \\
						& $S_m\uparrow\:$ & .755 & .768 & .794 & .740 & .781 & .791 & .809 & .713 & .798 & .803 & .815 & .802 & .819 & .823 & .823 & .818 & \textbf{.829} & \textbf{.829} \\
						& $F_\beta^w\uparrow\:$ & .661 & .668 & .696 & .633 & .684 & .722 & .704 & .626 & .722 & .751 & .739 & .720 & .754 & .743 & .761 & .752 & \textbf{.788} & {.77} \\
	\hline 
	\multirow{4}{*}{MB} & $M\downarrow\:$ & .138 & .128 & .115 & .160 & .105 & .113 & .099 & .139 & .111 & .106 & .121 & .109 & .104 & .100 & .101 & .100 & .09 & \textbf{.079} \\
						& $E_\xi^{m}\uparrow\:$ & .740 & .778 & .778 & .697 & .812 & .823 & .813 & .761 & .762 & .804 & .779 & .789 & .803 & .817 & .837 & .828 & .849 & \textbf{.862} \\
						& $S_m\uparrow\:$ & .685 & .720 & .742 & .657 & .762 & .744 & .775 & .696 & .734 & .754 & .751 & .762 & .764 & .792 & .777 & .774 & .785 & \textbf{.812} \\
						& $F_\beta^w\uparrow\:$ & .551 & .593 & .619 & .489 & .651 & .655 & .637 & .576 & .626 & .679 & .642 & .649 & .672 & .690 & .697 & .699 & .713 & \textbf{.732} \\
	\hline 
	\multirow{4}{*}{OC} & $M\downarrow\:$ & .149 & .130 & .129 & .168 & .119 & .116 & .119 & .169 & .109 & .115 & .119 & .121 & .112 & .111 & .105 & .106 & \textbf{.099} & {.103} \\
						& $E_\xi^{m}\uparrow\:$ & .755 & .784 & .780 & .706 & .800 & .808 & .784 & .718 & .820 & .810 & .801 & .798 & .809 & .800 & .829 & .817 & .83 & \textbf{.842} \\
						& $S_m\uparrow\:$ & .709 & .738 & .749 & .653 & .752 & .747 & .765 & .641 & .771 & .750 & .756 & .754 & .765 & .775 & .77 & .771 & .771 & \textbf{.778} \\
						& $F_\beta^w\uparrow\:$ & .593 & .622 & .630 & .520 & .644 & .659 & .638 & .527 & .680 & .672 & .659 & .658 & .678 & .673 & .687 & .683 & \textbf{.701} & {.692} \\
	\hline 
	\multirow{4}{*}{OV} & $M\downarrow\:$ & .184 & .159 & .150 & .216 & .136 & .125 & .127 & .217 & .129 & .134 & .148 & .146 & .119 & .126 & .113 & .120 & .124 & \textbf{.11} \\
						& $E_\xi^{m}\uparrow\:$ & .736 & .790 & .779 & .663 & .807 & .828 & .810 & .664 & .817 & .803 & .795 & .802 & .835 & .808 & .847 & .834 & .814 & \textbf{.855} \\
						& $S_m\uparrow\:$ & .688 & .728 & .745 & .624 & .751 & .762 & .781 & .611 & .761 & .748 & .747 & .752 & .779 & .774 & .791 & .779 & .764 & \textbf{.793} \\
						& $F_\beta^w\uparrow\:$ & .616 & .671 & .682 & .527 & .701 & .733 & .721 & .529 & .723 & .721 & .697 & .707 & .752 & .723 & \textbf{.761} & .749 & .732 & {.756} \\
	\hline 
	\multirow{4}{*}{SC} & $M\downarrow\:$ & .101 & .100 & .090 & .114 & .081 & .087 & .093 & .110 & .076 & .080 & {.075} & .083 & .078 & .078 & .076 & .080 & \textbf{.073} & .086 \\
						& $E_\xi^{m}\uparrow\:$ & .788 & .806 & .814 & .765 & .841 & .837 & .799 & .792 & .854 & .858 & .856 & .844 & .851 & .843 & \textbf{.878} & .860 & .871 & {.872} \\
						& $S_m\uparrow\:$ & .745 & .756 & .783 & .716 & .799 & .772 & .784 & .724 & .808 & .793 & .807 & .793 & .807 & {.809} & \textbf{.811} & .803 & \textbf{.811} & .808 \\
						& $F_\beta^w\uparrow\:$ & .593 & .611 & .638 & .550 & .677 & .669 & .627 & .594 & .696 & .708 & .695 & .678 & .706 & .691 & {.735} & .714 & \textbf{.739} & {.719} \\
	\hline 
	\multirow{4}{*}{SO} & $M\downarrow\:$ & .115 & .116 & .099 & .118 & .096 & .092 & .095 & .113 & .089 & .091 & .087 & .098 & .090 & .082 & {.08} & .087 & \textbf{.074} & {.08} \\
						& $E_\xi^{m}\uparrow\:$ & .747 & .752 & .769 & .732 & .780 & .802 & .766 & .759 & .792 & .804 & .814 & .784 & .801 & .797 & \textbf{.837} & .816 & .827 & \textbf{.837} \\
						& $S_m\uparrow\:$ & .703 & .706 & .737 & .682 & .732 & .736 & .748 & .682 & .746 & .745 & .768 & .749 & .755 & .767 & .772 &.763 & .775 & \textbf{.782} \\
						& $F_\beta^w\uparrow\:$ & .526 & .531 & .561 & .487 & .567 & .602 & .566 & .518 & .596 & .623 & .626 & .594 & .621 & .614 & .652 & .634 & \textbf{.671} & {.666} \\
	\hline 
\end{tabular}
\label{tab:soc}
\end{table*}
We follow recent methods to use the DUTS-TR (10553 images)\cite{DUTS} 
to train our network and resize images to 384x384. 
Normalization, random rotation, crop, and image enhancement (contrast, sharpness, brightness) 
were applied as the data augmentation. 
We set the batch size to 8 and the maximum epochs to 60. 
We use Adam optimizer\cite{kingma2014adam} to train our model 
with an initial learning rate 1e-4 (1/10 for the backbone). 
We use poly learning rate decay for scheduling with a factor of 
$(1-(\frac{iter}{iter_(max)})^{0.9})$ and linear warm-up for the first 12000 iterations. 
All experiments were implemented on an RTX 3090 GPU. 
\subsection{Evaluation Datasets and Metrics}
We evaluate our model on six widely used benchmark datasets, including DUT-OMORN\cite{DUT-O} (5168 images), 
DUTS-TE\cite{DUTS} (5019 images), ECSSD\cite{ECSSD} (1000 images), HKU-IS\cite{HKU-IS} (4447 images), 
PASCAL-S\cite{{PASCAL-S}} (850 images), SOD\cite{SOD} (300 images).

We employ four metrics to evaluate our model and the existing state-of-the-art methods:

\textbf{(1) Mean Absolute Error (MAE)}. MAE measures the average pixel-wise difference between the prediction $P$ and the ground truth $G$, and is calculated as $MAE = \frac{1}{H\times W}\sum_{x=1}^{H}\sum_{y=1}^{W}\vert P(x,y)-G(x,y)\vert$.


\textbf{(2) Enhanced-alignment measure ($E_\xi$)}. $E_\xi$\cite{ijcai2018p97} takes into account the local pixel values along with the image-level mean value. It is defined as $E_\xi=\frac{1}{H\times W}\sum_{x=1}^{H}\sum_{y=1}^{W}\phi_\xi(i,j)$, where $\phi_\xi$ represents the enhanced alignment matrix.

\textbf{(3) Structure-measure ($S_m$)}. $S_m$\cite{fan2017structure} aims to quantify the structural similarity between the prediction and the ground truth. It is calculated as $S_m = \alpha S_o +(1-\alpha)S_r$, where $S_o$ and $S_r$ correspond to the object-level and region-level structural similarity, respectively. The parameter $\alpha$ is set to 0.5.

\textbf{(4) Weighted F-measure ($F_\beta^w$)}. $F_\beta^w$\cite{6909433} provides a comprehensive extension of $F_\beta$\cite{Achanta2009FrequencytunedSR}. It is determined by the formula $F_\beta^w=\frac{(1+\beta^2)\times Precision^w\times Recall^w}{\beta^2\times Precision^w+Recall^w}$. This extension allows for real-valued TP, TN, FP, and FN, while assigning different weights ($w$) to errors based on their locations and neighborhood information. Notably, predictions with shadows or undersaturated regions yield lower $F_\beta^w$ scores. 
\subsection{Comparisons With State-of-the-Arts}
We compare the proposed method with 26 recent state-of-the-art CNN-based methods and 5 latest state-of-the-art Transformer-based methods,
including RAS\cite{RAS}, NLDF\cite{luo2017non}, C2SNet\cite{xin2018c2s}, SRM\cite{8237695}, R3Net\cite{R3Net}, 
BMPM\cite{Zhang_2018_CVPR}, DGRL\cite{Wang_2018_CVPR}, PiCANet\cite{PiCANet}, RANet\cite{8966594}, 
AFNet\cite{AFNet}, BASNet\cite{BASNet}, CPD\cite{CPD}, PoolNet\cite{PoolNet}, ITSD\cite{ITSD}, 
EGNet\cite{EGNet}, BANet\cite{Su_2019_ICCV}, SCRN\cite{SCRN}, F3Net\cite{F3Net}, 
TSPOANet\cite{TSPOANet}, MiNet\cite{MiNet}, LDF\cite{wei2020label}, CSF\cite{gao2020sod100k}, 
PFSNet\cite{ma2021pyramidal}, RCSB\cite{Ke_2022_WACV}, ICON\cite{ICON}, MENet\cite{wang2023pixels}, 
VST\cite{VST}, SRformer\cite{SelfReformer}, ISPRNet\cite{kim2022revisiting}, and BBRF\cite{BBRF}. 

\begin{figure*}
	\centering
	\includegraphics[scale=0.88]{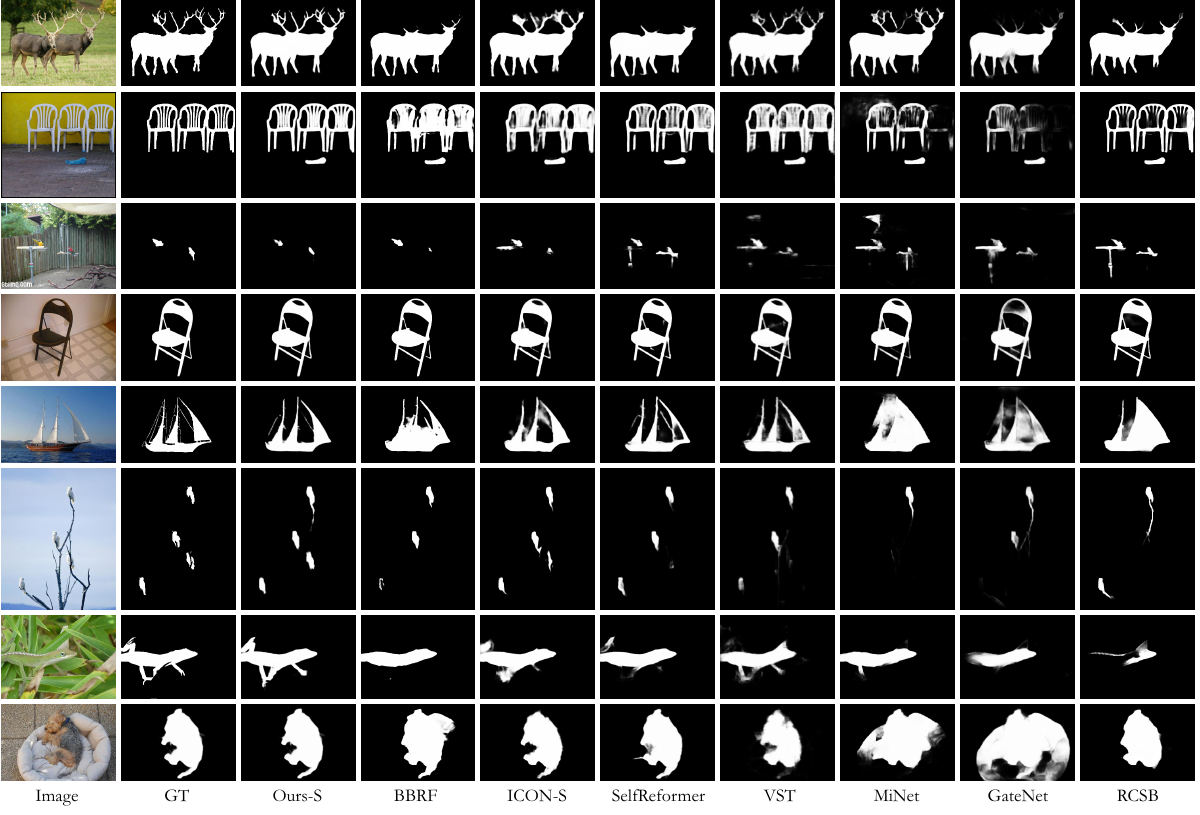}
	\vspace{-0.5em}\caption{Visual comparisons between proposed method and the other 7 state-of-the-art methods. 
	In contrast to previous methods, our method generates prediction maps with reduced shadows and undersaturated regions across various scenes. 
	}	

	\label{fig:visual}
\end{figure*}

\begin{table}
	\centering
	\scriptsize
	\caption{Ablation study of proposed modules. The best results are shown in \textbf{bold}. }
	\label{tab:ablation}
	\setlength\tabcolsep{0.5mm}
	\begin{tabular}{c|l|ccc|ccc|ccc}%
		\hline
		\multirow{2}{*}{ID} & \multirow{2}{*}{Component Settings} & \multicolumn{3}{c|}{DUTS} & \multicolumn{3}{c|}{ECSSD} & \multicolumn{3}{c}{HKU-IS}\\
		&& $M\downarrow$ & $S_m\uparrow$ & $F_\beta^w\uparrow$ & $M\downarrow$ & $S_m\uparrow$ & $F_\beta^w\uparrow$ & $M\downarrow$ & $S_m\uparrow$ & $F_\beta^w\uparrow$ \\
		\hline
		1 & Baseline & .046 & .866 & .776 & .047 & .904 & .871 & .037 & .907 & .866 \\
		2 & +MIA & .041 & .887 & .812 & .037 & .926 & .898 & .032 & .923 & .891 \\
		3 & +MIA+SSCA & .038 & .901 & .844 & .034 & .929 & .909 & .031 & .926 & .897 \\
		\rowcolor{gray!20} 4 & +MIA+SSCA+URA & \textbf{.033} & \textbf{.906} & \textbf{.865} & \textbf{.029} & \textbf{.934} & \textbf{.922} & \textbf{.025} & \textbf{.932} & \textbf{.917} \\
		\hline 
	\end{tabular}
\end{table}
\subsubsection{Quantitative Comparison}
Table \ref{tab:quantitative_comparison} shows the quantitative comparison results on six widely used benchmark datasets. 
We compare our method with 12 state-of-the-art methods in terms of MAE, $E_\xi$, $S_m$, $F_\beta^w$. 
The results show that our method outperforms all previous state-of-the-art methods. 
Notably, our method achieves convincing improvement 
in terms of the $F_\beta^w$\cite{6909433} metric compared to the previous best model, on six datasets. 
{Despite the integration of numerous attention mechanisms, our model maintains computational feasibility, enabling real-time inference speed (44 fps). }
In addition, we present the precision recall\cite{6871397} and F-measure curves\cite{Achanta2009FrequencytunedSR} in Figure \ref{fig:pr&fm}. 

Recently, Fan \textit{et al.} proposed a challenging SOC\cite{fan2022salient} dataset based on real-world scenarios. 
Compared to the previous six SOD datasets, this dataset includes more complex scenes and 
divides the dataset into nine different subsets based on the attributes of the images, 
including AC (appearance change), BO (big object), CL (clutter), HO (heterogeneous object), 
MB (motion blur), OC (occlusion), OV (out-of-view), SC (shape complexity), and SO (small object). 
We believe that evaluating the performance of our proposed model in comparison to previous methods 
on the SOC dataset can more comprehensively validate the performance of our model. 
In Table \ref{tab:soc}, we compare our model with 17 recent state-of-the-art 
CNN-based methods in terms of attribute-based performance. 
As can be seen, our model exhibits convincing performance improvement over the existing methods. 
Notably, our model achieved great performance improvement on the subset with the attribute of motion blur (MB). 
This indicates the effective handling of blur areas by our model. 
{Additionally, our method exhibits suboptimal performance on the subset with shape complexity (SC). We hypothesize that this could be attributed to the regional fragmentation caused by the ADP mechanism, which could have a particularly severe impact on objects with shape complexity. }

\subsubsection{Visual Comparison}
Figure \ref{fig:visual} provides visual comparisons between our method and other methods. 
As can be seen, our method reconstructs saliency maps with higher accuracy, 
effectively minimizing the presence of shadows and undersaturated regions. 
Besides, our method excels in dealing with challenging cases like small objects (rows 3 and 6), 
low-contrast (row 7), complex backgrounds (rows 3, 7, and 8), delicate structures (rows 1, 2, 4, and 5), 
and multiple objects (rows 2, 3 and 7) integrally and noiselessly. 
The above results show the versatility and robustness of our method. 

\subsection{Ablation Studies}

\begin{table}
	\centering
	\scriptsize
	\caption{Different feature enhancement methods compared with the proposed MIA module. }
	\label{tab:mia_vs_fem}
	\setlength\tabcolsep{0.6mm}
	\begin{tabular}{c|l|ccc|ccc|ccc}%
		\hline
		\multirow{2}{*}{ID} & \multirow{2}{*}{FEMs Settings} & \multicolumn{3}{c|}{DUTS} & \multicolumn{3}{c|}{ECSSD} & \multicolumn{3}{c}{HKU-IS}\\
		&& $M\downarrow$ & $S_m\uparrow$ & $F_\beta^w\uparrow$ & $M\downarrow$ & $S_m\uparrow$ & $F_\beta^w\uparrow$ & $M\downarrow$ & $S_m\uparrow$ & $F_\beta^w\uparrow$ \\
		\hline
		\rowcolor{gray!20} 2& +MIA & \textbf{.041} & .887 & \textbf{.812} & \textbf{.037} & \textbf{.926} & \textbf{.898} & \textbf{.032} & \textbf{.923} & \textbf{.891} \\
		5& +DFA\cite{ICON} & .045 & .878 & .796 & .04 & .92 & .892 & .035 & .915 & .88 \\
		6& +ASPP\cite{ASPP} & .045 & {.87} & {.783} & {.046} & {.908} & {.876} & {.037} & {.908} & {.868} \\
		7& +PSP\cite{zhao2017pspnet} & .044 & {.879} & {.797} & {.041} & {.919} & {.888} & {.036} & {.915} & {.877} \\
		8& +RFB\cite{Liu_2018_ECCV} & .043 & {.882} & {.807} & {.039} & {.921} & {.893} & {.035} & {.915} & {.88} \\
		9& +AIM\cite{MiNet} & .042 & \textbf{.888} & {.809} & \textbf{.037} & {.923} & \textbf{.898} & {.033} & {.919} & {.887} \\
		\hline 
	\end{tabular}
\end{table}
\begin{table}
	\centering
	\scriptsize
	\caption{Different interaction settings in proposed MIA module. }
	\label{tab:mia}
	\setlength\tabcolsep{0.4mm}
	\begin{tabular}{c|c|ccc|ccc|ccc}%
		\hline
		\multirow{2}{*}{ID} & \multirow{2}{*}{Interaction Settings} & \multicolumn{3}{c|}{DUTS} & \multicolumn{3}{c|}{ECSSD} & \multicolumn{3}{c}{HKU-IS}\\
		&& $M\downarrow$ & $S_m\uparrow$ & $F_\beta^w\uparrow$ & $M\downarrow$ & $S_m\uparrow$ & $F_\beta^w\uparrow$ & $M\downarrow$ & $S_m\uparrow$ & $F_\beta^w\uparrow$ \\
		\hline
		\rowcolor{gray!20} 2& $ 2 \backslash 3 \backslash 4 \backslash $ - & \textbf{.041} & \textbf{.887} & \textbf{.812} & \textbf{.037} & \textbf{.926} & \textbf{.898} & \textbf{.032} & \textbf{.923} & \textbf{.891} \\
		10& $ 3 \backslash 3 \backslash$-$\backslash $ - & .042 & .885 & .809 & .038 & .924 & .894 & .034 & .918 & .882 \\
		11& $ 3 \backslash 3 \backslash 4 \backslash $ - & .042 & \textbf{.887} & {.81} & \textbf{.037} & \textbf{.926} & {.897} & {.033} & \textbf{.923} & {.889} \\
		12& $ 4 \backslash 4 \backslash 4 \backslash $ - & .043 & {.884} & {.807} & {.038} & {.923} & {.895} & {.034} & {.92} & {.886} \\
		13& - $ \backslash 1 \backslash 2 \backslash 3 $ & .044 & {.882} & {.808} & {.041} & {.914} & {.886} & {.035} & {.913} & {.877} \\
		\hline 
	\end{tabular}
\end{table}
\subsubsection{Effectiveness of Different Components}

To demonstrate the effectiveness of the proposed modules, 
we conduct the quantitative results of several simplified versions of our method. 
We start from a UNet-like structure with skip connections as the baseline, 
and progressively incorporate the proposed modules, including MIA, SSCA, and URA. 
As can be seen in Table \ref{tab:ablation}, the model's performance demonstrates 
consistent improvement across metrics as the modules are incrementally incorporated. 

\begin{figure}
	\centering
\includegraphics[scale=0.88]{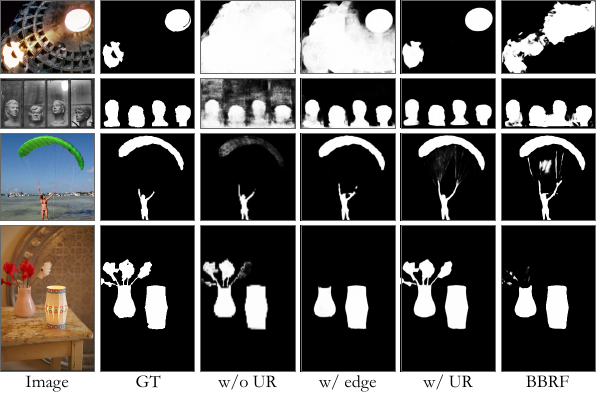}
\vspace{-1em}\caption{Visual demonstration for ablation study of proposed uncertainty refinement. 
	The results of BBRF were used for comparison. }
\label{fig:wo_ur}
\end{figure}

For the process of uncertainty refinement, visual comparisons are provided in Fig \ref{fig:wo_ur}. 
These visual comparisons demonstrate the effectiveness of our uncertainty refinement 
in eliminating shadows and undersaturated regions within saliency predictions, 
resulting in high-quality fine-grained saliency predictions.

\subsubsection{MIA vs. Feature Enhancement Methods}
To verify the superiority of the proposed MIA, 
we compared it with existing feature enhancement methods, 
including DFA\cite{ICON}, ASPP\cite{ASPP}, PSP\cite{zhao2017pspnet}, AIM\cite{MiNet}, 
and RFB\cite{Liu_2018_ECCV}. As can be seen in Table \ref{tab:mia_vs_fem}, 
the proposed MIA demonstrates advantages across all performance metrics. 
This noticeable improvement can be attributed to the inherent limitation of existing feature enhancement methods, wherein uniform processing is applied to features of varying scales, disregarding their distinctive properties. In contrast, the proposed MIA effectively capitalizes on the complementarity between features at different scales. 
Furthermore, compared to AIM\cite{MiNet}, the proposed MIA leverages interactive attention to establish perception without the need for feature interpolation, thereby avoiding information loss during feature interaction and achieving superior performance. 

We further conducted experiments on different interconnections between multilevel features, 
and the result is shown in Table \ref{tab:mia}. 
As described in \ref{text:MIA}, the proposed MIA incorporates interactions 
between the high-level feature and the current-level feature. 
To calculate the interaction attention for F$_1$, F$_2$, and F$_3$, we utilize F$_2$, F$_3$, and F$_4$, respectively, 
shown as ``$2\backslash3\backslash4\backslash-$" in Table \ref{tab:mia}. 
The experiments conducted on different interaction linkages provide the following insights: 
1. Cross-scale perception proves effective not only for shallow features but also for deep features 
(ID: 10, 11). 
2. Perceiving neighboring scales leads to enhanced performance (ID: 2, 11, and 12). 
3. Interconnections from higher layers to lower layers outperform the interconnections 
from lower layers to higher layers (ID: 2, 13). 

\begin{table}
	\centering
	\scriptsize
	\caption{Different saliency integration methods compared with the proposed SSCA module. }
	\label{tab:ssca_vs_sims}
	\setlength\tabcolsep{0.3mm}
	\begin{tabular}{c|l|ccc|ccc|ccc}%
		\hline
		\multirow{2}{*}{ID} & \multirow{2}{*}{Integration Methods} & \multicolumn{3}{c|}{DUTS} & \multicolumn{3}{c|}{ECSSD} & \multicolumn{3}{c}{HKU-IS}\\
		&& $M\downarrow$ & $S_m\uparrow$ & $F_\beta^w\uparrow$ & $M\downarrow$ & $S_m\uparrow$ & $F_\beta^w\uparrow$ & $M\downarrow$ & $S_m\uparrow$ & $F_\beta^w\uparrow$ \\
		\hline
		\rowcolor{gray!20} 3 & +MIA+SSCA & {.038} & \textbf{.901} & \textbf{.844} & \textbf{.034} & \textbf{.929} & \textbf{.909} & \textbf{.031} & \textbf{.926} & \textbf{.897} \\
		14 & +MIA+SIM\cite{MiNet} & \textbf{.037} & {.895} & {.834} & {.035} & {.925} & {.904} & {.032} & {.921} & {.893} \\
		15 & +MIA+AFM\cite{ma2021pyramidal} & .04 & {.894} & {.837} & {.036} & {.923} & {.901} & \textbf{.031} & {.923} & {.894} \\

		\hline 
	\end{tabular}
\end{table}

\subsubsection{SSCA vs. Saliency Integration Methods} 
In order to demonstrate the superior performance of the proposed SSCA 
in saliency integration, we conduct a comparative evaluation with existing feature aggregation methods, 
including SIM\cite{MiNet}, AFM\cite{ma2021pyramidal}, shown in Table \ref{tab:ssca_vs_sims}. 
{Previous feature aggregation methods also focus on integrating salient information; however, they remain constrained by the limited receptive field of convolutional operations, which hinders their ability to effectively model salient objects at a global level. }Conversely, the proposed SSCA capitalizes on the attention mechanism to holistically model salient objects, facilitating the integration of consistent and integral saliency information. As a result, enhanced performance is achieved. 

{In this work, we are the first to explicitly define and distinguish between feature enhancement methods (FEMs) and feature aggregation methods (FAMs). We define feature enhancement as the process of refining and improving the raw features extracted by the feature extractor. The goal is to prepare these features for subsequent aggregation by suppressing noise, amplifying salient information, and addressing inconsistencies across different feature levels. Techniques for feature enhancement may include, but are not limited to, diverse convolutional operations~\cite{ICON}, channel attention~\cite{tracer}, and spatial attention~\cite{kim2021uacanet} mechanisms. Feature aggregation involves integrating features from different levels to consolidate salient information into a unified representation. This process aims to produce a coherent and complete saliency map. In this work, we refer to this process as ``Saliency Integration" to emphasize its purpose.
We believe that distinguishing between FEMs and FAMs is meaningful, as it helps to clarify the feature enhancement-to-aggregation pipeline, thereby enabling us to explore performance improvements in saliency detection from multiple perspectives. Note that, this pipeline has already been adopted by numerous methods~\cite{MiNet,kim2022revisiting,ma2021pyramidal}, and our work serves to explicitly clarify and formalize it. }
\subsubsection{Uncertainty Guidance vs. Boundary Guidance}
\begin{table}
	\centering
	\scriptsize
	\caption{Different guidance settings in our URA. 
	}
	\label{tab:ura}
	\setlength\tabcolsep{0.4mm}
	\begin{tabular}{c|c|ccc|ccc|ccc}%
		\hline
		\multirow{2}{*}{ID} & \multirow{2}{*}{Guidance Settings} & \multicolumn{3}{c|}{DUTS} & \multicolumn{3}{c|}{ECSSD} & \multicolumn{3}{c}{HKU-IS}\\
		&& $M\downarrow$ & $S_m\uparrow$ & $F_\beta^w\uparrow$ & $M\downarrow$ & $S_m\uparrow$ & $F_\beta^w\uparrow$ & $M\downarrow$ & $S_m\uparrow$ & $F_\beta^w\uparrow$ \\
		\hline
		16& None & .04 & .901 & .862 & .031 & .933 & .921 & .027 & .93 & .912 \\
		17& Boundary & .039 & .898 & .853 & .031 & .933 & .920 & .026 & {.931} & .914 \\
		\rowcolor{gray!20} 4& Uncertainty & \textbf{.033} & \textbf{.906} & \textbf{.865} & \textbf{.029} & \textbf{.934} & \textbf{.922} & \textbf{.025} & \textbf{.932} & \textbf{.917} \\
		\hline 
	\end{tabular}
\end{table}
We compare boundary guidance with uncertainty guidance in the proposed URA. 
In the control group labeled as ``none", we do not mask the attention calculation. 
In the ``boundary" group, we adopt the boundary extraction method from EGNet\cite{EGNet} to 
extract the boundary information from the input image, 
which then serves as guiding information for the URA module. 
As can be seen in Table \ref{tab:ura}, compared to no guidance (ID: 16),
boundary guidance (ID: 17) resulted in only a negligible improvement in the metrics, 
with slight declines observed in certain metrics. 
Conversely, uncertainty guidance (ID: 4) demonstrated a more substantial improvement in the metrics. 
{An additional qualitative comparison is presented in Fig \ref{fig:wo_ur}. }
We attribute this {performance gap} to the ability of uncertainty guidance to highlight under-saturated regions in the predicted saliency map, whereas boundary guidance relies on fixed prior knowledge and cannot be adjusted according to the model's prediction. These experimental results confirm our hypothesis. 

\subsubsection{Effectiveness of Adaptive Dynamic Partition}
We compared the proposed Adaptive Dynamic Partition (ADP) mechanism with two other partition methods: small partition (ID: 18) and random partition (ID: 19). 
Specifically, the small partition involves dividing the feature map into uniformly sized windows, 
akin to window partition\cite{Swin}. 
Meanwhile, in random partition, we assign a partition probability of 50\%. 
As shown in Table \ref{tab:adp}, the metrics achieved by the small partition and random partition were lower compared to the proposed ADP. This disparity is likely a result of the excessive fragmentation of uncertainty regions associated with these partition approaches. Such fragmentation may disrupt the continuity of contextual information, thereby hindering the model's perception of uncertainty. However, the proposed ADP selectively refrains from partitioning regions characterized by a high degree of uncertainty, effectively mitigating fragmentation and preserving the integrity of these regions. 
Moreover, we conducted an evaluation of the performance of the ADP mechanism at different $p_t$ (ID: 20, 21). 
From the results in Table \ref{tab:adp}, 
it can be seen that the proposed ADP can better serve the role of uncertainty guidance. 

\begin{table}
	\centering
	\scriptsize
	\caption{Different partition settings in our URA. }
	\label{tab:adp}
	\setlength\tabcolsep{0.5mm}
	\begin{tabular}{c|c|ccc|ccc|ccc}%
		\hline
		\multirow{2}{*}{ID} & \multirow{2}{*}{Partition Settings} & \multicolumn{3}{c|}{DUTS} & \multicolumn{3}{c|}{ECSSD} & \multicolumn{3}{c}{HKU-IS}\\
		&& $M\downarrow$ & $S_m\uparrow$ & $F_\beta^w\uparrow$ & $M\downarrow$ & $S_m\uparrow$ & $F_\beta^w\uparrow$ & $M\downarrow$ & $S_m\uparrow$ & $F_\beta^w\uparrow$ \\
		\hline
		18& Small Partition & .04 & {.9} & {.857} & {.03} & {.928} & {.916} & {.026} & {.924} & {.91} \\
		19& Rand Partition & .038 & {.901} & {.854} & {.031} & {.926} & \textbf{.922} & {.027} & {.925} & {.911} \\
		\rowcolor{gray!20} 4& ADP and $p_t$ = 0.2 & \textbf{.033} & \textbf{.906} & \textbf{.865} & \textbf{.029} & \textbf{.934} & \textbf{.922} & \textbf{.025} & \textbf{.932} & \textbf{.917} \\
		20& ADP and $p_t$ = 0.1 & .035 & {.902} & {.86} & {.03} & {.932} & {.921} & {.026} & {.928} & {.913} \\
		21& ADP and $p_t$ = 0.4 & .037 & {.9} & {.857} & {.031} & {.929} & {.92} & {.027} & {.926} & {.912} \\
		\hline 
	\end{tabular}
\end{table}

\subsubsection{Evaluation of Loss Function}
Previous methods commonly incorporate weights in the loss function 
to attach higher importance to challenging regions, 
directing the model's attention toward these areas during training. 
F$^3$Net\cite{F3Net} was the first to introduce this weighting mechanism and proposed 
$\mathcal{L}_{wbce}$ and $\mathcal{L}_{wiou}$ to train the network. 
TRACER\cite{tracer} further introduced a multi-scale mechanism in the weighting 
and proposed $\mathcal{L}_{api}$, which can adaptively weight based on pixel relationships. 
We also made efforts to enhance the proposed model's focus 
on uncertain regions using this weighting strategy. 
Unfortunately, this approach did not result in the expected improvement in model performance. 
As presented in Table \ref{tab:loss}, contrary to expectations, 
training with weighted loss (ID: 23, 25) led to suboptimal performance. 
We suppose our proposed method already offers sufficient perception guidance to the model, 
and the introduction of weighted loss biases the focus toward complex regions, 
thereby hindering the model's attention to non-complex regions. 
\begin{table}
	\centering
	\scriptsize
	\caption{Different loss function settings with the proposed network. 
	}
	\label{tab:loss}
	\setlength\tabcolsep{0.3mm}
	\begin{tabular}{c|c|ccc|ccc|ccc}%
		\hline
		\multirow{2}{*}{ID} & \multirow{2}{*}{Loss Settings} & \multicolumn{3}{c|}{DUTS} & \multicolumn{3}{c|}{ECSSD} & \multicolumn{3}{c}{HKU-IS}\\
		&& $M\downarrow$ & $S_m\uparrow$ & $F_\beta^w\uparrow$ & $M\downarrow$ & $S_m\uparrow$ & $F_\beta^w\uparrow$ & $M\downarrow$ & $S_m\uparrow$ & $F_\beta^w\uparrow$ \\
		\hline
		22& $\mathcal{L}_{bce}$ & .039 & {.901} & {.841} & {.032} & {.932} & {.91} & {.029} & {.93} & {.903} \\
		23& $\mathcal{L}_{api}$\cite{tracer} & {.035} & {.897} & {.859} & {.029} & {.927} & {.918} & {.027} & {.925} & {.911} \\
		24& $\mathcal{L}_{bce} + \mathcal{L}_{iou}$ & .036 & {.903} & {.862} & \textbf{.028} & \textbf{.934} & \textbf{.922} & {.026} & {.93} & {.914} \\
		25& $\mathcal{L}_{wbce} + \mathcal{L}_{wiou}$\cite{F3Net} & .042 & {.899} & {.852} & {.03} & {.933} & {.921} & {.027} & {.928} & {.911} \\
		\rowcolor{gray!20} 4&$\mathcal{L}_{bce} + \mathcal{L}_{iou} + \mathcal{L}_{sc}$ & \textbf{.033} & \textbf{.906} & \textbf{.865} & {.029} & \textbf{.934} & \textbf{.922} & \textbf{.025} & \textbf{.932} & \textbf{.917} \\
		\hline 
	\end{tabular}
\end{table}
\subsection{Limitations}
While our proposed model effectively addresses shadows 
and under-saturated regions in the prediction map, 
resulting in high-quality refined saliency predictions, 
it is important to acknowledge some limitations. 
Specifically, the generation of the uncertainty map relies on the saliency map 
and is unable to rectify erroneous localization of salient objects. 
Additionally, the flexibility of window partition is constrained and may 
lead to the fragmentation of uncertain regions in specific scenarios {(e.g., objects with shape complexity)}. 
We remain committed to researching solutions to mitigate these shortcomings. 

\section{Conclusion}
In this paper, we propose an uncertainty guidance learning approach 
for fine-grained salient object detection (SOD) to address the issues of shadows 
and under-saturated regions frequently occurring in saliency prediction maps. 
Firstly, we alleviate the limitations of existing aggregation methods by 
introducing the Multilevel Interaction Attention (MIA) module and the Scale Spatial-Consistent Attention (SSCA) module, extracting high-quality salient information. 
Then, we design the Uncertainty Refinement Attention (URA) module to effectively eliminate artifacts and under-saturated regions in the prediction maps. 
The URA module leverages uncertainty guidance, improving the model's perception of 
uncertain areas and ultimately generating highly-saturated fine-grained saliency prediction. 
Additionally, we introduce the Adaptive Dynamic Partition (ADP) mechanism to strike a balance between model performance and computational cost. {Notably, our refinement approach also can be seamlessly integrated into other methods, showing promising potential for application to other binary image segmentation tasks. }
Finally, we present a novel network, dubbed UGRAN, 
that integrates the above components. 
Experiments demonstrate the superior performance of our proposed network on seven widely used datasets. 

\section*{Acknowledgments}
The authors would like to thank the anonymous reviewers and editor for their valuable comments, which have significantly improved the quality of this manuscript. 

\bibliographystyle{IEEEtran}
\bibliography{references}

\begin{IEEEbiography}[{\includegraphics[width=1in,height=1.25in,clip,keepaspectratio]{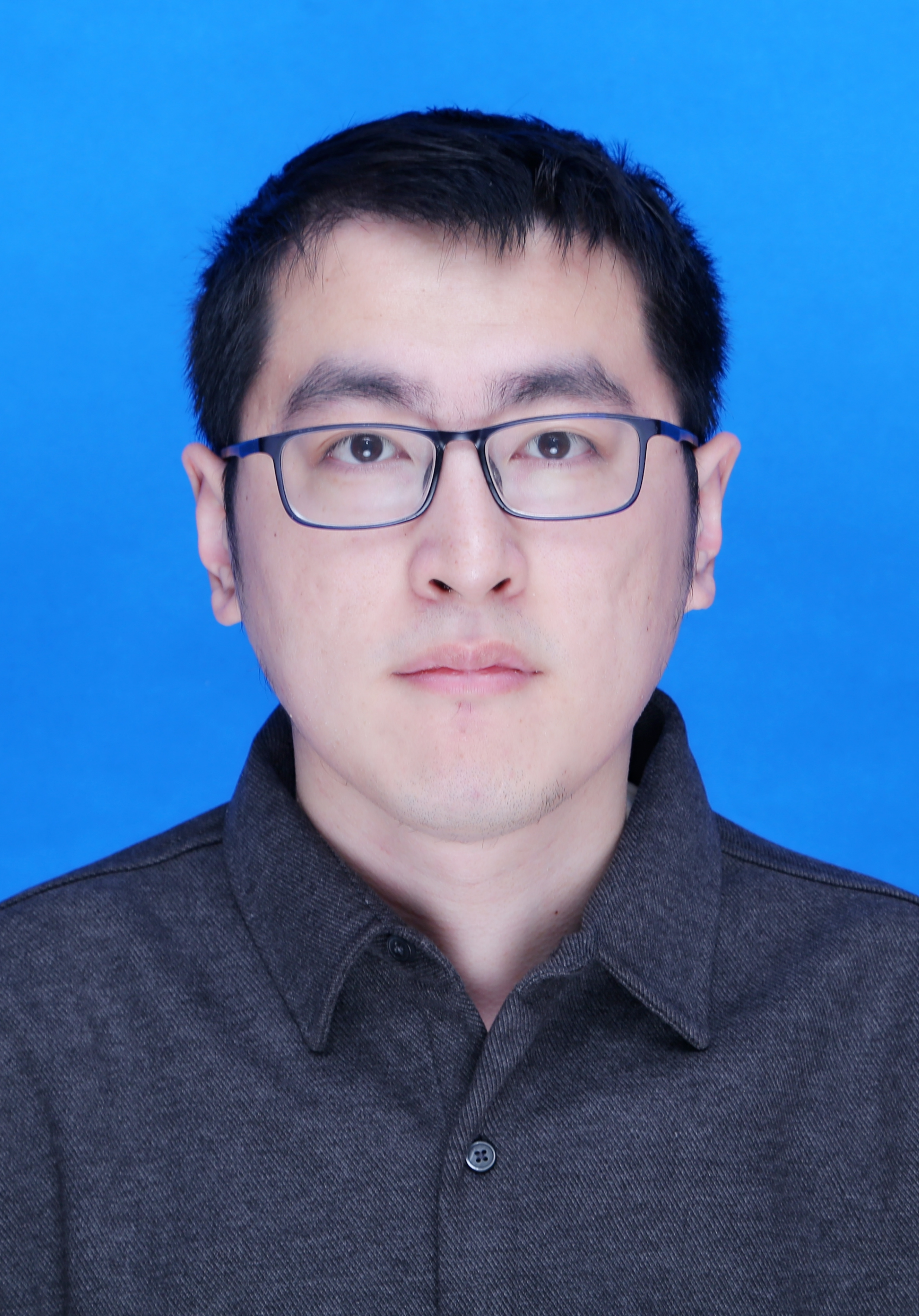}}]{Yao Yuan}
received the B.S. degree in Computer Science and Technology from Jiangsu University of Science and Technology, Zhenjiang, China, in 2021. He is currently working toward the M.S. degree in Electronic Information with the Department of Computer Science and Technology, Nanjing University of Aeronautics and Astronautic, Nanjing, China. His research interests include Unified Salient Object Detection and Fine-Grained Prediction.
\end{IEEEbiography}

\begin{IEEEbiography}[{\includegraphics[width=1in,height=1.25in,clip,keepaspectratio]{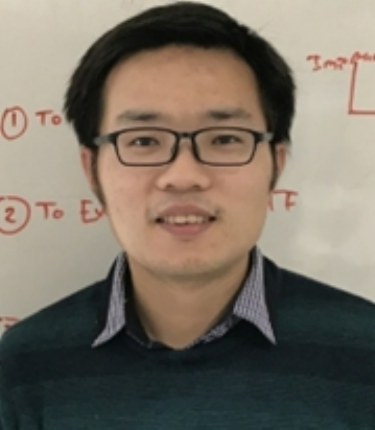}}]{Pan Gao}
received the Ph.D. degree in electronic engineering from University of Southern Queensland (USQ), Toowoomba, Australia, in 2017. Since 2016, he has been with the College of Computer Science and Technology, Nanjing University of Aeronautics and Astronautics, Nanjing, China, where he is currently an Associate Professor. He has authored or co-authored more than 80 publications in scientific journals and international conferences. His research interests include deep learning, computer vision, and artificial intelligence.

\end{IEEEbiography}

\begin{IEEEbiography}[{\includegraphics[width=1in,height=1.25in,clip,keepaspectratio]{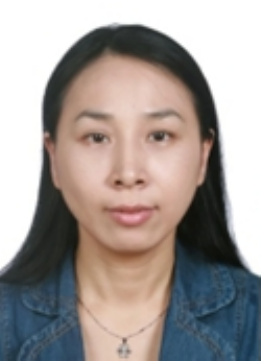}}]{Qun Dai}
became an Associate Professor at the College of Computer Science and Technology, Nanjing University of Aeronautics and Astronautics (NUAA), in 2010. Since 2015, she has become a Professor at the same university. Her research interests focus on neural computing, pattern recognition, and machine learning.

\end{IEEEbiography}

\begin{IEEEbiography}[{\includegraphics[width=1in,height=1.25in,clip,keepaspectratio]{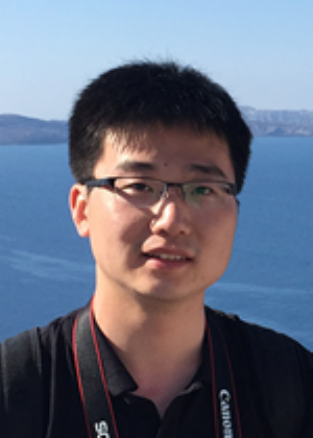}}]{Jie Qin}
is currently a Professor at Nanjing University of Aeronautics and Astronautics, China. He received the B.E. and Ph.D. degrees from Beihang University, China, in 2011 and 2017, respectively. His current research interests include computer vision and machine learning. He has published over 90 papers in top-tier journals/conferences, including IEEE TPAMI, IJCV, CVPR, ICCV, ECCV, AAAI, IJCAI, and NeurIPS. He is serving/served as a Guest Editor of IJCV, a Program Chair of an ECCV Workshop, Senior PC members of AAAI and IJCAI, and Area Chairs of ACM MM, ECAI, IJCB, and PRCV.
\end{IEEEbiography}

\begin{IEEEbiography}[{\includegraphics[width=1in,height=1.25in,clip,keepaspectratio]{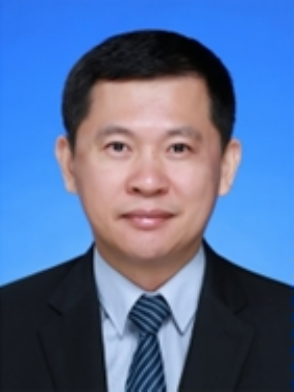}}]{Wei Xiang}
is Cisco Research Chair of AI and IoT and Director of the Cisco-La Trobe Centre for AI and IoT at La Trobe University. Previously, he was Foundation Chair and Head of Discipline of IoT Engineering at James Cook University, Cairns, Australia. Due to his instrumental leadership in establishing Australia’s first accredited Internet of Things Engineering degree program, he was inducted into Pearcy Foundation’s Hall of Fame in October 2018. He is a TEDx speaker and an elected Fellow of the IET in UK and Engineers Australia. He received the TNQ Innovation Award in 2016, and Pearcey Entrepreneurship Award in 2017, and Engineers Australia Cairns Engineer of the Year in 2017. He was a co-recipient of four Best Paper Awards at WiSATS’2019, WCSP’2015, IEEE WCNC’2011, and ICWMC’2009. He has been awarded several prestigious fellowship titles. He was named a Queensland International Fellow (2010-2011), an Endeavour Research Fellow (2012-2013), a Smart Futures Fellow (2012-2015), and a JSPS Invitational Fellow (2014-2015).  He is currently an Associate Editor for IEEE Communications Surveys \& Tutorials, IEEE Transactions on Vehicular Technology, IEEE Internet of Things Journal, etc. He has published over 300  papers including 3 books and 220 journal articles. His research interest includes the Internet of Things, wireless communications, machine learning for IoT data analytics, and computer vision. 

\end{IEEEbiography}
	
\end{document}